# Deep-Mobility: A Deep Learning Approach for an Efficient and Reliable 5G Handover

Rahul Arun Paropkari, Anurag Thantharate, Cory Beard
*School of Computing and Engineering*
*University of Missouri - Kansas City, Missouri, USA*
{rap8vb, adtmv7, beardc}@umkc.edu

*Abstract* – 5G cellular networks are being deployed all over the world and this architecture supports ultra-dense network (UDN) deployment. Small cells have a very important role in providing 5G connectivity to the end users. Exponential increases in devices, data and network demands make it mandatory for the service providers to manage handovers better, to cater to the services that a user desire. In contrast to any traditional handover improvement scheme, we develop a 'Deep-Mobility' model by implementing a deep learning neural network (DLNN) to manage network mobility, utilizing in-network deep learning and prediction. We use network key performance indicators (KPIs) to train our model to analyze network traffic and handover requirements. In this method, RF signal conditions are continuously observed and tracked using deep learning neural networks such as the Recurrent neural network (RNN) or Long Short-Term Memory network (LSTM) and system level inputs are also considered in conjunction, to take a collective decision for a handover. We can study multiple parameters and interactions between system events along with the user mobility, which would then trigger a handoff in any given scenario. Here, we show the fundamental modeling approach and demonstrate usefulness of our model while investigating impacts and sensitivities of certain KPIs from the user equipment (UE) and network side.

*Keywords – 5G Cellular Handover, Machine Learning, Deep Learning Neural Network, Handover Optimization, Vertical Handover, HetNets, Fade Duration Outage Probability.*

I. INTRODUCTION

The ever-growing increase in the number of devices and services has led the mobile industry to grow exponentially in the past few years. Connected devices in future cellular networks are not only limited to mobile handsets and tablets, but belong to a wide range of equipment coming from an overlapping area of eMBB (enhanced mobile broadband), mMTC (massive machine type communication) and URLLC (ultra-reliable low latency communication). The data demands on these devices are huge and the bandwidth requirements are in Gbps due to the video-on-demand services offered today. 4G-LTE was able to cater to most needs supporting 720p on video but most content available going forward will be Full HD, UHD offering 4K and 8K video quality. Augmented Reality (AR), Virtual Reality (VR) also add to the bandwidth demands on some online shopping applications to provide a rich shopping experience. According to Ericsson's Mobility report from November 2018, the global mobile data traffic is expected to increase eight times between 2017–2023 and video content about five times being offered on over 70% of whole mobile data traffic [3] triggering the eMBB services.

Service providers are forced to provide an infrastructure for serving various applications, use cases, and business with superior Quality of Service (QoS). In some industries like the public safety, emergency, medical, or energy sectors, mobile devices provide great capabilities but must be highly reliable. In order to provide uninterrupted services to these devices on the wireless network, current 4G-LTE infrastructure has struggled over time and has been just about up to the mark meeting high reliability standards, very low latency and highspeed user connectivity for URLLC.

Another pool of devices, in the verge of explosion, are the IoT (internet of things) devices catering to the mMTC services of the 5G networks transforming the home, office, city streets, public places, and beyond. These all have a specific QoS requirement and are treated differently in handover (HO) scenarios, if any. Until now, mobility was one of the prime factors to trigger a handover in any given network but several other parameters like Received Signal Strength (RSSI), Signal to Interference and Noise Ratio (SINR), Fade Duration (FD), Quality of Service (QoS), backhaul connectivity, network congestion, reliability, etc., may also result in a handover. We propose a scheme that will accommodate such parameters to make a combined/intelligent decision towards handover by capturing the multi-dynamic nature of a handover.

II. PROBLEM STATEMENT AND NEXT GENERATION CELLULAR HANDOVERS

Network controlled and UE assisted hard HO procedures are adopted in 3GPP LTE-Advanced. Handover Margin (HOM), Time-To-Trigger (TTT) and A3offset are the major parameters based on which the entering condition of A3event is initiated. The existing HO mechanisms are damaging the 5G latency and availability requirements as about 70% of the current cellular HO are still depending on the RSRP (signal power) and/or the RSRQ (signal quality) as a single parameter considered to trigger a handover. With recent advancements, the management of cellular handovers has become more complex. Networks have become more heterogeneous (HetNets) with multiple technologies achieving the same goal of providing endless connectivity to a user. With the on-going increase in the complexity of the network, manual tuning is very time consuming and very much more prone to errors than before. In addition, RF and channel conditions may change due to several environmental and surrounding factors. Human intervention at every critical step isn't enough, so SONs are gaining importance.

Service providers are implementing some of the SON use cases such as the Mobility Robustness Optimization (MRO) introduced in 3GPP Release 9 [4]. However, the conventional MRO algorithm tunes handover parameters based on counts of handover failures and handover events; and only makes use of the hysteresis to improve HO performance. Unlike in the past, multiple network parameters must be tuned and synchronized to achieve certain goals. 3GPP had also proposed the Automatic Neighbor Relation (ANR) tool for early LTE deployments which is still widely used in the industry. Many HO research areas propose modifying the HOM and TTT in order to make handover decisions precise and accurate for different environments. Achieving load balancing has been an important goal of many academic research areas, and service providers implemented similar solutions like Wi-Fi Offloading, traffic balance using the Femto Access Points (FAP), etc.

Handovers need to keep up a balance in order to avoid the ping pong effect and also too much loading of any one specific cell. With the current and future networks consisting of massive small cell deployment, especially to overcome the uncertainty of millimeter wave (mm-wave) communications, it is important to minimize call drops during HO and avoid redundant and unnecessary HO. A small yet temporary obstacle between a UE and any mm-wave gNB can impact connectivity and trigger a HO even without UE mobility. The increasing probability of HOs may cause HO failure (HOF) or HO ping-pong which degrades the system performance. Use of SONs are to raise these automation standards and provide the required balance.

If more SONs are running with the same goal on different objects, then there's a conflict of interest and that network parameter can be altered wrongly. Certain cell (C-1) might be overloaded and needs some users moved to a neighbor (C-2) and SON-1 takes appropriate action and moves 20% users from C-1 to C-2 by adjusting power levels of C-2. At the same time, SON-2 is making sure all users get a RSRP of a certain value which C-1 can provide at this time and not C-2. So, it's trying to move users back to C-1 causing an overload and ping pong effect.

The main reason for adjusting the same parameter is that not all parameters are equally attractive from the optimization viewpoint. For example, some 3GPP mobility parameters, such as the TTT and the Hysteresis, are not defined on a per cell-basis or adjacency-basis, which is a severe limitation in mobile networks due to the variant nature of the radio environment. On the contrary, the HOM is a mobility parameter defined per adjacency-basis that provides greater optimization capabilities. In short, multiple parameters need to be considered at the same time to make an educated decision. Fig. 1 shows the basic handover message exchange which can either be initiated by the UE or by the network. Fig. 2 shows HO requirements such as the HOM or the Hysteresis which is the minimum threshold RSRP value between the serving cell and the target cell power levels. If this value is reached, network begins to count the TTT and if this value is maintained for a minimum of pre-defined TTT, a HO will be triggered. All the above mentioned justifies the need to involve more UE and network related parameters and network-based triggers towards a HO decision model.

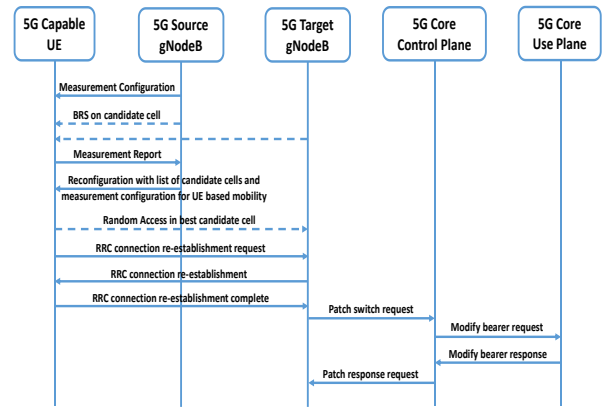

Figure 1. UE/Network Initiated Handover Procedure

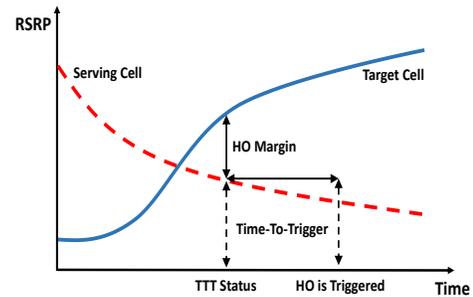

Figure 2. Handover Triggering Decision making Parameters

Currently, the base station neighbor list is populated using the ANR functionality of the SONs, but not on a real-time basis. Service providers are looking for a non-static and dynamic way of doing this on instantaneous basis. So, we have a list of neighbors, but then additional parameters are required to figure out which one to connect to if need be. Rather than only considering power levels, HO decision can be made using multiple other parameters like the bandwidth capacity of a gNB, current user load, backhaul capacity, future maintenance activities in the database, RF channel stability, modulation and coding schemes (MCS), dual connectivity (DC), etc.

We create a dataset that contains some information from the UE and some from other network resources. Our dataset consists of the UE type used, supported technology, time and day of the connection, RSRP, RSRQ of serving as well as some (3-4) neighboring cell sites, RF channel conditions based on MCS, available channel bandwidth, backhaul capacity, alarm status on cell sites, maintenance tickets if any, etc. Our DLNN learns from this information and will help identify the following: Too many HOs and when/where, neighbor list needs to be updated, power levels of each eNB at every location, RF conditions, behavior of the UE, movement and velocity of the UE, routine mobility patterns of the UE, power levels of the eNB in UL, UE bandwidth demands, applications being used and other service parameters the from network and RF perspective. Since service providers are deploying dense networks using pico, femto, UE relays, DAS systems, cellular and Wi-Fi hotspots, they are working harder to find a trade-off between cost, coverage and complexity of network and handover management.

## III. RELATED WORK

There has been significant work done on Handover Optimization using various different techniques. As for the best of our knowledge, our work is unique as it is the first to assume multiple parameters from UE and network side to come together to trigger a mobility-based handover decision using a DLNN technique. An eNB pre-selection strategy is proposed in [1] so that high-speed user equipment is primarily taken care by macro eNBs while low-speed UEs or UEs with low quality of service (QoS) requirements are offloaded to the service area of pico eNBs to share the load of macro eNBs. The authors in [2] make use of reinforcement learning (RL) to control the handovers between base-stations using a centralized RL agent. This agent handles the radio measurement reports from the UEs and chooses appropriate handover actions in accordance with the RL framework to maximize a long-term utility. Ericsson mobility report predicts the growth of mobile devices, 5G network connections and the overall data usage in coming years [3]. To overcome handover latency, [5] jointly considers edge and core delays, with a novel cost-effective software-defined ultra-dense framework by dynamically removing state execution times.

The authors in [6] have a game theoretical approach implemented and evaluated for dense small cell heterogeneous networks to validate the enhancement achieved in the proposed method. As for network intelligence, the authors in [7] represented handovers using matrix exponential distributions for public safety and emergency communications, which helps make handover decisions more accurate considering all the different parameters involved in the decision process. To solve the unnecessary HOs and ping-pong, [8] proposes a weighted fuzzy self-optimization (WFSO) approach for the optimization of the handover control parameters (HCPs) considering SINR, traffic load of serving and target BS, and UE velocity. The 5G Network Slicing concept is fully utilized to manage the network traffic and route the connections to the most appropriate slice using DLNNs and understanding of what the connection demands in [9]. Multiple network and RF parameters are considered before making such an intelligent decision using neural networks. In [10] authors propose a scheme to control some parameters related to the A3 trigger event for handover management in order to deploy mobility load balancing (MLB) and MRO independently. They modify the Cell Individual Offsets (CIO) and adjust the Hysteresis and TTT autonomously.

Authors in [11] contrast Fade Duration Outage Probability (FDOP) based handover requirements with the traditional SINR based handovers methods in cellular systems. The research in [12] evaluates the performance parameter of X2 based handover on one network provider in Cirebon area and optimizes handover parameters using RSRP algorithm and RSRQ algorithm. In order to accomplish successful HO, authors in [13] study the impact of HOM, A3offset along with TTT and extend their analysis to different distances from BS for varying UE velocity. Authors in [14] present their ongoing work and vision towards automating end-to-end model management in deep learning. We used their prototype to effectively track all our experiments and also perform detailed analysis. Authors present system-level architectural changes on both UE and Network elements along with a proposal to modify control signaling as part of Radio Resource Control messages using smartphone battery level in [15]. Authors in [16] present a network slicing based architecture that also includes a quarantine slice to isolate suspicious flows in the network and protects against various types of denial-of-service and other attacks.

## IV. PROPOSED SOLUTION MODELS - RNN AND LSTM

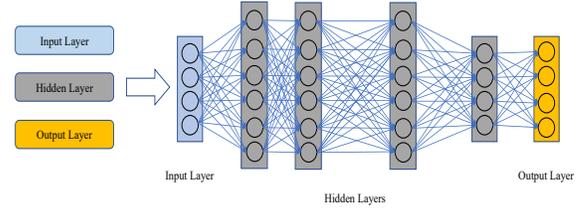

Figure 3. General Deep Learning Neural Network

In today's day and age, prediction has become very intuitive and common on most platforms. However, the prediction problem (and algorithms) are still one of the hardest in data science industry including a wide variety, as in predicting sales, stock markets, speech recognition, sequential prediction on what you will type next while searching, or even some sort of guess work by Alexa/Google/Siri assistants, etc. With NNs being in buzz today, and also practically implemented in many forms, the use of LSTMs has proven to be the most effective solution so far. Recent studies have proven some other less computational and yet equally effective solutions, but LSTMs stay in business. Here we consider RNNs and LSTMs, for reasons listed below.

### A. Recurrent Neural Networks (RNNs)

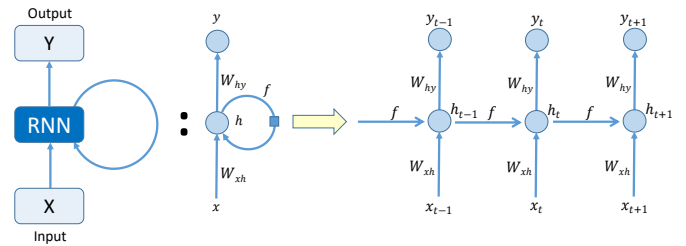

Figure 4. A typical Recurrent Neural Network Model

In any conventional feed forward network, all variables are treated independently, meaning any prediction done will not take into account previous output results or any data from a day before. Therefore, RNNs come in play to achieve the time dependency. Fig. 4, shows what a typical RNN will look like and what it mathematically means when expanded. RNNs will take into consideration an event from the immediate previous state and will let the present operation be influenced from earlier results. RNNs have several limitations, and for a continuous time dependent string of data, they tend to not completely understand the context behind an input. They are good, especially when dealing with short term dependencies and universal inputs. So, to go back in the past and relate to something from that far, to help make a decision today, is out of scope of the plain RNN networks and so LSTMs were introduced.

## B. Long Short-Term Memory Neural Networks (LSTM)

RNNs often tend to fail in extracting the most appropriate context from the long data feed and this is typically directly related to the vanishing gradient problem in neural networks. When we look at how a network learns, any weights applied are actually a cross product of the learning rate, error from earlier layers and the input to this particular layer. The previous layer error is a product of all previous errors which are theoretically small values. When an activation function is applied to any of the layers, for example a sigmoid function, even smaller values of the derivatives of errors get multiplied several times, resulting in an infinitesimal small value propagated backwards to the earlier layers in the network. And thus, the gradient almost vanishes as we go back toward the earlier layers in the network. This is what happens in standard RNNs where immediate information is available for a short period, but with large datasets, LSTMs are the way to achieve accuracy and reliability.

Unlike the RNNs which apply a single function to transform the whole information, LSTMs follow a slightly different approach of splitting up the information into relevant and non-relevant ones. As shown below in Fig. 5, this is achieved with some simple math functions of additions and multiplications, which allows for more control over the flow and mixing of inputs as per trained weights. In addition, LSTM information flows through 'state', this state is the horizontal line on the top that allows for basic mathematical function to alter the output state. Standard activation functions like the sigmoid (output between 1 and 0) and tanh (output between +1 and -1) are also used to take limited information to the next states.

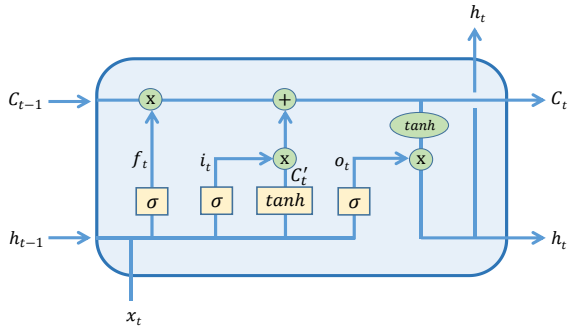

Figure 5. Standard Long Short-Term Memory Neural Network

The LSTMs have the ability to remove or add information to the cell state, carefully regulated by structures called gates. The first and the left most is the 'forget gate' (output denoted by $f_t$), which applies a sigmoid $\sigma$ function to previous $h_{t-1}$ and $x_t$ to allow whatever information to take in, 1 meaning take all and a 0 forgets all. The second, and central portion, is a combination of two small functions: first is the input gate in conjunction with a sigmoid function (output denoted by $i_t$), and second is the $tanh$ function to create a new vector $C'_t$ that gets added to the state above. This will give us the new state $C_t$ based on the applied functions and the previous state $C_{t-1}$ as shown in below derivations. Finally, we get to our output which will be based on our cell state, but will be a filtered version (denoted by $h_t$). First, we run a sigmoid layer which decides what parts of the cell state we're going to output (denoted by $o_t$). Then, we put the cell state through $tanh$ and multiply it by the output of the sigmoid gate, so that we only output the parts we decided to. All the derived equations for LSTM are shown below.

$$f_t = \sigma(x_t U^f + h_{t-1} W^f)$$
$$i_t = \sigma(x_t U^i + h_{t-1} W^i)$$
$$C'_t = tanh(x_t U^g + h_{t-1} W^g)$$
$$C_t = \sigma(f_t * C_{t-1} + i_t C'_t)$$
$$o_t = \sigma(x_t U^o + h_{t-1} W^o)$$
$$h_t = tanh(C_t) * o_t$$

## V. PROPOSED LEARNING MODEL - DEEPMOBILITY

The current cellular network 4G-LTE architecture has a rigid framework and often lacks customization when it comes to offering any tailored business requirements. Service providers have to often rely on third party vendors or equipment manufacturers to provide desirable customer solutions. SONs come into the picture to introduce automatic adaptability in the network like the ANR and MRO functionality does the neighbor list updates and modifications to the hysteresis and TTT values for any targeted base station or a whole cluster at once. However, this change is not performed based on the dynamic network or RF changes but based on some service providers' database information from collective HO performances as in HO failures (HOF), call drop rates (CDR), link failure rates (LRF), etc. The current RF conditions might have changed between now and the times these rates were recorded. Some network parameters are modified by engineers during a maintenance window overnight.

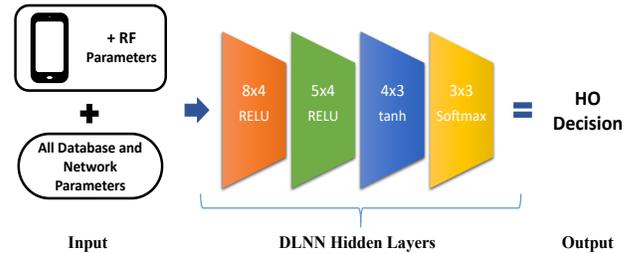

Figure 6. Our Deep-Mobility Neural Network Model

Our model consists of an input layer with number of neurons same as the input parameters, a couple of hidden layers and an output layer. Each layer is activated using a certain activation function to make parameters non-linear and close to the real world. A 4x3 means a total of 12 neurons in that particular layer. Output layer is applied a simple linear regression function. About 30% dataset was used for validation. Even though the major factor for handovers is movement, there is not one single factor in a real-world scenario which triggers a handover. Handover dynamics come from a combination of simple network aspects like the Received Signal Strength Indicator (RSSI), Signal to Interference and Noise Ratio (SINR), Fade Duration (FD), Quality of Service (QoS), shadowing, network congestion, resource availability of a given network, etc. Some other dynamic factors causing a handover are multipath, Doppler spread, or velocity. Even the slightest change in the location in a dense cellular network may cause a user to handover to a neighboring cell. For example, if SNR is the only handoff trigger factor, then maybe we have an acceptable SNR in an area, but the radio resources available are not enough and so we get low throughput in spite of a good SNR. If there's an overlapping cell in that region which can provide greater throughput at a slightly lower SNR then the user needs to be handed off to this new cell.

## VI. RESULTS

We created a whole dataset of variables from real network deployment for all serving and neighboring base stations which partly contributes to about 65% of our input dataset used. This UE measurement report information contains the power (RSRP) of serving as well as three to four neighboring cell sites. We make use of multiple third-party mobile applications to retrieve this information; up to four different applications were used to compare and contrast the base stations' data. We also identify these base stations as a 3G (BTS), 4G-LTE (eNodeB) or a 5G base station (gNodeB). Base station identification could be easily done based on the frequency band and the bandwidth reported from these cell sites using the mobile application. Table I below shows you some of the basic parameters that are captured by the UE and these Apps for all sites that the device can communicate with. All this was done using multiple commercial 5G and LTE based cell phone devices/UEs.

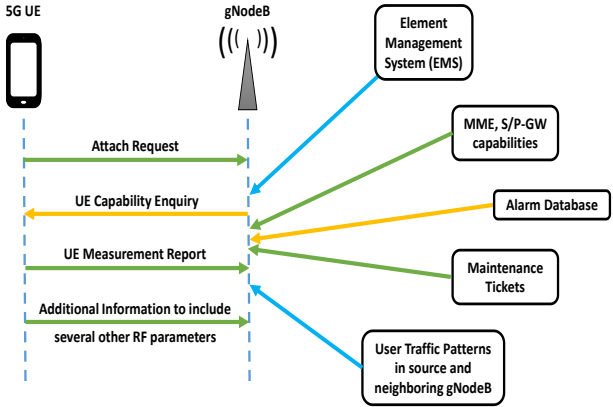

Figure 7. Feature Highlights of UE Measurement Report and Network Centric Parameters

| MCC | Mobile Country Code | EARFCN | E-UTRA Absolute RF Channel Number |
|---|---|---|---|
| MNC | Mobile Network Code | RSRP | Reference Signal Received Power |
| TAC | Tracking Area Code | RSRQ | Reference Signal Received Quality |
| eNBID | eNodeB Identification | RSSI | Received Signal Strength Indicator |
| PCI | Physical Cell Identity | CQI | Channel Quality Indicator |
| CID | Cell Identity | SID | System Identification |
| BAND | Cellular Band Used | NID | Network Identification |

Table I. Major KPIs Reported by the Mobile Applications

As seen above, applications provide this information with each object having a unique value at every physical location the UE will collect this data from. Most of them are standard and self-explanatory, TAC typically is used to point the BS to a pre-defined pool of Mobile Management Entity (MMEs) and often limited to a geographic area. CID is a unique identification code to determine each sector of each band and is determined as per the service providers' naming scheme. EARFCN is an ETSI industry standard to display the channel numbers instead of raw frequencies in MHz and is bandwidth independent. This is an important parameter in our training process to analyze when was the carrier aggregation (CA) is enabled and under what conditions it was not available. We have other RF parameters which can be captured either by the UE or on the base stations.

In our case, we made some assumptions on these variables for the purpose of training our DLNN. Most of them cannot be accurately determined, unless captured over a long duration of time, as in the dynamic RF channel conditions, signal modulation used on the downlink (DL) and the uplink (UL) for each cell BS, scheduling of the Resource Blocks (RBs), etc. We have some of this information captured and shown in Table II for understanding purposes. We would assume some of these neighboring sites would have some future maintenance tickets open and are reflected in the database system. We also have a real time monitoring performed by the Element Management System (EMS) which will have present alarms reported from the BS. SONs will be pre-programmed to take action on certain regular alarms but we don't make that assumption here and consider manual intervention will be necessary to fix any alarming issues. Some alarms can be fixed remotely and some require dispatching of a crew to the cell site. Our dataset is not concerned with the alarm fixing process but will only categorize if an existing alarm is serving impacting or not. We flag all serving impacting alarms in our input dataset so our DLNN can differentiate and learn their impacts on users.

| System Entity | Parameter | Description |
|---|---|---|
| EMS | Alarm 1 | Allowed |
| | Alarm 2 | Service Impacting |
| Maintenance Database | Ticket 1 | Allowed |
| | Ticket 2 | Service Impacting |
| KPI Capturing Tool | CFR | Call Failure Rate |
| | CDR | Call Drop Rate |
| | HOF | Handover Failure Rate |
| | RLF | Radio Link Failure Rate |
| | Cell Load | Number of connected users |

Table II. Additional network side attributes in HO decision

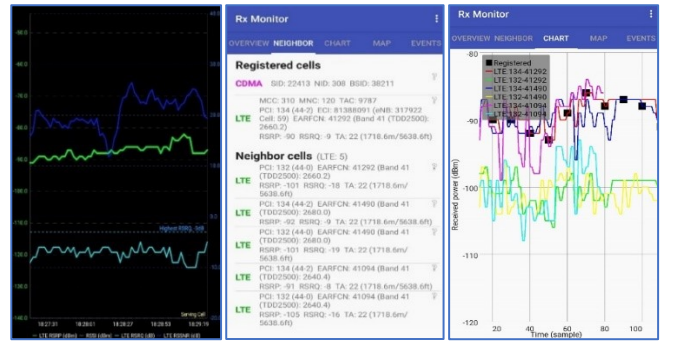

Figure 8 (a-b-c). Sample snapshot of application reports

If we assume a UE-K in our network and in connected mode. Our 'Deep-Mobility' model has learned about the movement patterns of this UE-K for the past few weeks and understands that for this particular time, UE-K will be pretty much in a half mile radius of the current reported location. Assume that better RSRP and RSRQ conditions are reported for the $\alpha$ (alpha) sector of a target eNB-T than an existing $\beta$ (beta) sector of the serving eNB-S. Then the network or the UE would initiate a handoff request to eNB-T. What if the EMS has reported a service impacting alarm on that $\beta$ sector of eNB-T which the UE is not aware about, and the network, even when aware, would otherwise not consider this during a HO decision. The UE is better off by not handing over to the eNB-T as it seems to be immobile at this time and location based on the system learnings.

Figures 8 [a, b, c] show glimpses of what we could capture on the mobile applications regarding the RSRP, RSRQ and RSSI. It also shows to what technology is the UE connected to. Our dataset includes the most relevant KPIs from both the network and the device side, including the type of device connected, QoS Class Identifier (QCI), packet delay budget, maximum packet loss, time and day of the week, etc. These KPIs can be captured from control packets between the UE and network. Since our model will run internally on the network, all this information is readily available. Base on the training and testing we can plot below in Fig. 9 [a, b] the accuracy and loss for each of these activities. Accuracy of close to one means a very efficient training. Loss was minimum. Training of our 'Deep-Mobility' model involved the entire dataset and validation used 30% random data from the original input dataset.

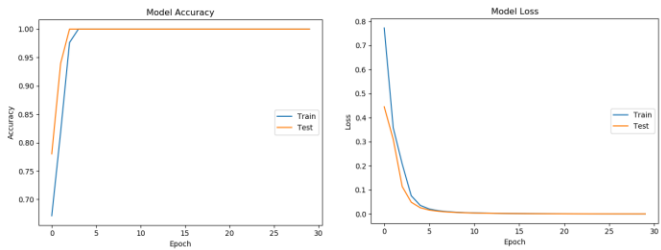

Figure 9 (a-b). Model accuracy and Loss for training and validation

## VII. CONCLUSION

The next generation of wireless networks will have multiple cell sites, and HO decisions will be very critical to reliable connections and sufficient QoS. Mobility management will be of utmost importance for a service provider and so will be the user experience, and use of Deep Learning to make intelligent network decisions will be a key in shaping the management of future autonomous and self-sufficient network. We designed a novel base model to learn and analyze the UE as well as the network parameters together and make the most optimal decisions in terms of the user mobility. We accommodate just as many variables as required to accurately predict handovers based on new input parameters fed by the UE and the network. Our proposed model is straightforward and flexible, yet captures all the detailed information for the parameters causing a cellular handoff. This will be especially useful for formulating the next generation of 5G/6G handover algorithms that will need to incorporate hyper-network densification, mmWave, M2M traffic, and ultra-reliable low latency communication.

## REFERENCES

[1] H. Ferng and Y. Huang, "Handover scheme with enode-B pre-selection and parameter self-optimization for LTE-A heterogeneous networks," *2016 International Conference on Machine Learning and Cybernetics (ICMLC)*, Jeju, 2016, pp. 594-599.

[2] Yajnanarayana, V., Rydén, H., Hévizi, L., Jauhari, A. and Cirkic, M., Apr 2019, "5G Handover using Reinforcement Learning" *arXiv preprint arXiv:1904.02572*.

[3] Ericsson 2018 Report - https://www.ericsson.com/assets/local/mobility-report/documents/2018/ericsson-mobility-report-november-2018.pdf

[4] 3GPP TS 36.300, Evolved Universal Terrestrial Radio Access (E-UTRA) and Evolved Universal Terrestrial Radio Access Network (E-UTRAN); Overall description ; Stage 2 (Release 9), 9.5.0 ed., 2010.

[5] M. Erel-Özçevik and B. Canberk, "Road to 5G Reduced-Latency: A Software Defined Handover Model for eMBB Services," in *IEEE Transactions on Vehicular Technology*, vol. 68, no. 8, pp. 8133-8144, Aug. 2019.

[6] M. Alhabo, L. Zhang, N. Nawaz and H. Al-Kashoash, "Game theoretic handover optimisation for dense small cells heterogeneous networks," in *IET Communications*, vol. 13, no. 15, pp. 2395-2402, 17-9-2019.

[7] R. A. Paropkari, C. Beard, A. V. De Liefvoort, "Handover performance prioritization for public safety and emergency networks," *2017 IEEE 38th Sarnoff Symposium*, Newark, NJ, USA, 2017, pp. 1-6.

[8] A. Alhammadi, M. Roslee, M. Y. Alias, I. Shayea, S. Alriah and A. B. Abas, "Advanced Handover Self-optimization Approach for 4G/5G HetNets Using Weighted Fuzzy Logic Control" *2019 15th International Conference on Telecommunications (ConTEL)*, Austria, 2019, pp. 1-6.

[9] A. Thantharate, R. A. Paropkari, C. Beard and V Walunj "DeepSlice: A Deep Learning Approach towards an Efficient and Reliable Network Slicing in 5G Networks," *2019 IEEE 10th Annual Ubiquitous Computing, Electronics & Mobile Communication Conference (UEMCON)*, Newyork, NY, USA, 2019, pp. 0762-0767.

[10] T. Bag, S. Garg, D. Preciado, Z. Shaik, J. Mueckenheim and A. Mitschele-Thiel, "Self-Organizing Network functions for handover optimization in LTE Cellular networks," *Mobile Communication - Technologies and Applications; 24. ITG-Symposium*, Osnabrueck, Germany, 2019, pp. 1-7.

[11] R. A. Paropkari, A. A. Gebremichail and C. Beard, "Fractional Packet Duplication and Fade Duration Outage Probability Analysis for Handover Enhancement in 5G Cellular Networks," *2019 International Conference on Computing, Networking and Communications (ICNC)*, Honolulu, HI, USA, 2019, pp. 298-302.

[12] S. L. Harja and Hendrawan, "Evaluation and optimization handover parameter based X2 in LTE network," *2017 3rd International Conference on Wireless and Telematics (ICWT)*, Palembang, 2017, pp. 175-180.

[13] A. S. Priyadharshini and P. T. V. Bhuvaneswari, "A study on handover parameter optimization in LTE-A networks," *2016 International Conference on Microelectronics, Computing and Communications (MicroCom)*, Durgapur, 2016, pp. 1-5.

[14] G. Gharibi, V. Walunj, S. Rella and Y. Lee, "ModelKB: Towards Automated Management of the Modeling Lifecycle in Deep Learning," 2019 IEEE/ACM 7th International Workshop on Realizing Artificial Intelligence Synergies in Software Engineering (RAISE), Montreal, QC, Canada, 2019, pp. 28-34, doi: 10.1109/RAISE.2019.00013.

[15] A. Thantharate, C. Beard and S. Marupaduga "An Approach to Optimize Device Power Performance Towards Energy Efficient Next Generation 5G Networks," *2019 IEEE 10th Annual Ubiquitous Computing, Electronics & Mobile Communication Conference (UEMCON)*, New York, NY, USA, 2019, pp. 0749-0754.

[16] A. Thantharate, R. Paropkari, V. Walunj, C. Beard and P. Kankariya, "Secure5G: A Deep Learning Framework Towards a Secure Network Slicing in 5G and Beyond," 2020 10th Annual Computing and Communication Workshop and Conference (CCWC), Las Vegas, NV, USA, 2020, pp. 0852-0857, doi: 10.1109/CCWC47524.2020.9031158.

[17] DeepSlice Dataset: https://github.com/adtmv7/DeepSlice (Last accessed date 09.25.2019) KPI's used for training and testing the model.